\documentclass{article} 
\usepackage{nips15submit_e,times}
\usepackage{url}
\usepackage[pdftex]{graphicx}
\usepackage{caption}
\usepackage{subcaption}
\usepackage{graphicx}
\usepackage{xspace}
\newcommand{\etal}{{et al.\@\xspace}}

\title{Empirical Study on Deep Learning Models for QA}

\author{
Yang Yu \quad Wei Zhang \quad Chung-Wei Hang \quad Bing Xiang \quad Bowen Zhou\\
        {IBM Watson}\\
       {\{yu, zhangwei, hangc, bingxia, zhou\}@us.ibm.com}\\
}

\nipsfinalcopy %

\begin{document}

\maketitle
\begin{abstract}
In this paper we explore deep learning models with memory component or attention mechanism for question answering task. We combine and compare three models, Neural Machine Translation \cite{NeuralMachineTranslation}, Neural Turing Machine \cite{NTM}, and Memory Networks \cite{MemNN} for a simulated QA data set \cite{TowardsAI-Complete}. This paper is the first one that uses Neural Machine Translation and Neural Turing Machines for solving QA tasks. Our results suggest that the combination of attention and memory have potential to solve certain QA problem.

\end{abstract}

\section{Introduction}
Question Answering (QA) is a natural language processing (NLP) task that requires deep understanding of semantic abstraction and reasoning over facts that are relevant to a question \cite{teachingMachines}. There are many different approaches to QA: constructing NLP pipeline where each component is separately trained and then assembled \cite{buildingWatson}, building large knowledge bases (KBs) \cite{SimpleQAwithMemNet} and reasoning with facts therein, and machine ``reading'' approach to comprehend question and documents \cite{teachingMachines} where answers are contained. Recently, various deep learning (DL) models are proposed for different learning problems. DL models are usually differentiable from end to end through gradient descent. They require neither any hand craft features nor separately tuned components. Thus we think it is important to study these models on addressing QA problem.

Implicitly or explicitly, solving QA problem can be divided into two steps. The first step locates the information that is relevant to the question, e.g. sentences in a text document or facts in a knowledge graph. We call it the ``search step.'' The second step which we call ``generation step'', extracts or generates answer from the relevant pieces of information detected in the search step. This paper focuses on reading comprehension type QA, where search and generation sometimes are coupled.

We focus on Neural Machine Translation (NMT), Memory Network (MemNN), and Neural Turing Machine (NTM) models, as they have representative state-of-the-art DL model architectures in categories they fall in. We conducted empirical studies in this work to better understand the strength and places to improve in each model on solving QA and experiment settings follow the 2-step QA framework. We will briefly describe the NMT, NTM and MemNN in the next section.

\section{Deep Learning Models for Question Answering}
\label{sec_model}
\textbf{MemNN}
MemNNs have been applied to QA \cite{MemNN, endtoend} and have shown promising results with different input transformation or model changes. Its strength mainly lies in reasoning with inference components combined with a long-term memory component and learning how to use these jointly. A general MemNN has four modules: \textit {input} which converts the incoming input to the internal feature representation, \textit{output} which produces a new output, \textit{generalize} which updates old memories given the new input, and \textit{response} which converts the output into the response format desired. The memory network described in \cite{MemNN} memorizes each fact in a memory slot and uses supporting facts for a given question labeled in the training data to learn  to search facts. Sainbayar {\etal}~\cite{endtoend} described another version of the MemNN that can be trained end-to-end without knowing the supporting facts.

\begin{figure}[t!]
 \begin{subfigure}[b]{0.5\textwidth}
        \centering
        \includegraphics[scale=0.3]{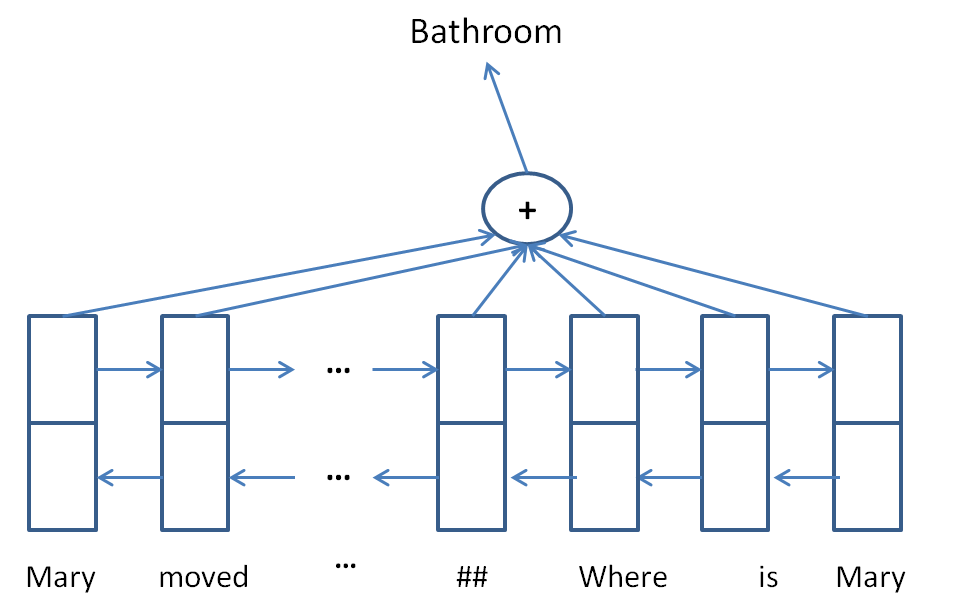}
        \caption{NMT model}
        \label{fig_nmt_model}
 \end{subfigure}
 \begin{subfigure}[b]{0.5\textwidth}
        \centering
        \includegraphics[scale=0.4]{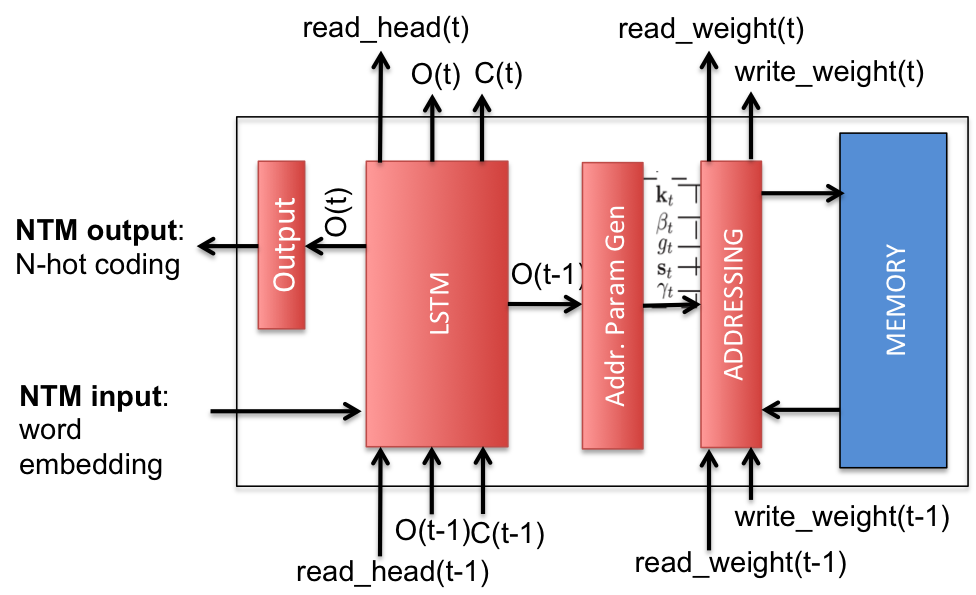}
        \caption{NTM model}
        \label{fig_ntm_model}
 \end{subfigure}
\caption{NTM and NMT models. }
\label{fig_nmt_ntm_model}
\end{figure}

\textbf{NMT}
Using Machine Translation (MT) technique for QA is not new, as generating answer by given question can be regarded as generating target text by given source text, or in other words, the answer is a translation of the question. Several previous works have used translation models to determine answers \cite{Cui:2005:QAP,Surdeanu:2011:LRA}. NMT brings new approaches to machine translation, for example two recurrent neural network (RNN) models are proposed \cite{recurrentContinuousTM,SequencetoSequenceLearningwithNN}. Following previous success of applying MT for QA, we think it is important to study if NMT could further help QA. As to the best of our knowledge, no one has done this study yet. Traditional translation systems usually are phrase-based where small sub-components are tuned separately (e.g., Koehn {\etal} \cite{StatisticalPhrase-basedTranslation}). NMT improves over phrase-based systems by using a neural network to encode a sentence in source language and decode it into a sentence in target language, which is a end-to-end system. However a main constraint of this encoder-decoder approach is that the model needs to compress all information of a source sentence into a fixed-length vector, which is difficult for long sentences or passages for reading comprehension style QA. In order to address this issue, Bahdanau {\etal} \cite{NeuralMachineTranslation} introduced an extension to the encoder-decoder model which uses bi-directional RNN and learns to align and translate jointly.
The NMT model we use is shown in the Figure \ref{fig_nmt_model}. The input includes passage and question and they are delimited by a marker. From the figure, we see the two RNNs read word by word in the input of different directions.
Each time when the model generates an answer word, it searches for multiple positions in the passage where the most relevant information is concentrated. The model then predicts an answer word based on the context vectors associated with these positions and all the previous generated answer words.
Formally, the $i$-th answer word $y_{i}$ (equation \ref{eq_nmt}) is conditioned on the words in answer before word $y_{i}$ and the passage $\mathbf{x}$. In RNN, the conditional probability is modeled as a nonlinear function $g()$ which depends on previous answer word $y_{i-1}$, the RNN hidden state $s_{i}$ for time $i$ and the context $c_{i}$. The context vector $c_i$ is a weighted sum of a sequence of annotations. Each annotation has information about the complete passage with a focus on the parts around the $i$-th word.
\begin{equation}
\label{eq_nmt}
p(y_{i}|y_1,...,y_{i-1},\mathbf{x})=g(y_{i-1},s_{i},c_{i})=g(y_{i-1},f(s_{i-1},y_{i-1},c_{i}),c_{i})
\end{equation}

\textbf{NTM} NTM \cite{NTM} resembles Turing machines in that it could learn arbitrary procedure in theory. As we believe that QA problem can be solved with (probably sophisticated) programs, in this paper we would examine how well NTM performs on reading comprehension tasks. NTMs are essentially RNNs, which in turn are Turing-complete \cite{Siegelmann95} and capable of encoding any computer program in theory, yet not always practical. NTM is end-to-end trainable with gradient descent, due to the fact that every component is differentiable that is enabled by converting the hard read and write into `blurry' operations that interact to greater or lesser degree. The read and write heads, as well as the memory component are recurrently updated through time, no matter if the controller is recurrent or not. This paper is the first to examine Neural Turing Machines on QA problems.
Our implementation of NTM (see Figure \ref{fig_ntm_model}) internally uses a single-layer LSTM network as controller. NTM inputs are word distributed representations \cite{w2v}. Word embedding is directed to LSTM controller within NTM, and the output of NTM is generated from softmax layer where each bit of output corresponds to an answer. In doing so we regard the QA problem as a multi-class (multi-word) classification problem given that the number of answers is finite and small. When multiple answers are needed, top n words with top n probabilities will be selected.

\section{Experiments}
The AI-Complete data \cite{TowardsAI-Complete} is a synthetic dataset designed for a set of AI tasks. For each task, small and large training sets have $1k$ and $10k$ questions, respectively, and test set has $1k$ questions. The supervision in the training set is given by the true answers to questions, and the set of supporting facts for answering a given question. All tasks are so clean that a human could achieve 100\% accuracy, as they are generated with a simulation. In this data, every statement in a passage has a number ID and each question is associated with the true answer and the IDs of supporting facts.

\begin{table}
\scriptsize
\caption{AI-complete 20 tasks results.}
\label{tbl_results}
\centering
\resizebox{\textwidth}{!}{
\begin{tabular}{|l|c|c|c|c|c|c|c|c|c|c|}
\hline
&\multicolumn{3}{c|}{(i) Memory Network}&\multicolumn{2}{c|}{(ii) Support fact only}&\multicolumn{2}{c|}{(iii) Sup. fact highlighted}&\multicolumn{1}{c|}{(iv) Combination }&\multicolumn{2}{c|}{(v) No support fact}\\
\hline
&a&b&c&d&e&f&g&h&i&j\\
Task&MemNN&MemNN-S&MemNN-R&NMT&NTM&NMT&NTM&MemNN-S+NMT&NMT&NMT(10k)\\
\hline
1- 1 Supporting Fact&100&100&100&100&100&100&100&100&98.2&99.6\\
2- 2 Supporting Facts&100&66.4&100&100&100&99.6&100&69.7&41.3&98.4\\
3- 3 Supporting Facts&100&49.3&100&100&100&99.5&100&60.4&33.4&97.1\\
4- Two Arg. Relations&69&66.9&100&99.1&100&97.5&100&66.5&97.8&99.8\\
5- Three Arg. Relations&83&98.5&82.5&99.3&79.2&90.6&73.7&98.1&90.3&91.7\\
6- Yes/No Questions&52&100&49.4&100&100&99.8&100&100&84.6&78.9\\
7- Counting&78&58.3&69.5&98.5&100&96.6&100&44.8&82.4&93.7\\
8- Lists/Sets&90&N/A&N/A &99&100&92.7&98&N/A &70.8&55.2\\
9- Simple Negation&71&100&63.6&100&100&99.7&100&100&89.3&86.6\\
10- Indef. Knowledge&57&100&49.3&98.9&94.6&96.8&85.9&98.9&73.5&79.5\\
11- Basic Coreference &100&79.5&55.5&100&100&100&100&84.4&99.8&97.9\\
12- Conjunction&100&100&100&100&100&100&100&100&99.4&98.9\\
13- Comp. Coreference &100&89&49.5&100&100&100&100&91.7&99.7&99.1\\
14- Time Reasoning&100&99.9&100&99.8&100&97.5&100&99.8&44.4&96.7\\
15- Basic Deduction&73&100&100&100&100&92.7&100&100&42.9&96\\
16- Basic Induction&100&92.8&100&100&100&88.1&100&96.1&42.7&52.7\\
17- Pos. Reasoning&46&100&46.7&64.2&69.3&58&61.2&56.9&64.6&89.5\\
18- Size Reasoning&50&27.2&52.9&97.8&93&91.8&93&58.3&90.9&96.5\\
19- Path Finding&9&30.7&24.6&80.7&100&29.7&100&16.9&9.3&44.2\\
20- Agent’s Motivations&100&100&100&100&100&93.3&100&100&91.6&91.7\\
\hline
Mean Performance&78.9&82.0&76.0&96.9&96.7&91.2&95.6&81.2&72.3&87.2\\
\hline
\end{tabular}}
\end{table}

MemNN \cite{MemNN} has shown promising results on AI-Complete data. So first we want to learn how different MemNN components contribute to it. We choose the MemNN with adaptive memory \cite{TowardsAI-Complete} as baseline (column a). We divide it into two virtual steps as the 2-step QA framework. We believe searching supporting facts (MemNN-S) is very important to MemNN while the MemNN response module (MemNN-R) has more room to improve, because it does not have any attention mechanism to focus on words in retrieved facts. We train MemNN-S by using complete passage as input and supporting facts as prediction and train MemNN-R by using true supporting facts as input and final answer as prediction. Column b,c in Table \ref{tbl_results} confirm our hypothesis that the accuracy of searching supporting facts from MemNN-S is much better than the accuracy of predicting final answer words from MemNN-R.

Following our findings in the first experiment, we want to see if NMT or NTM could do better than MemNN-R. Using the same setting as testing MemNN-R, both NMT and NTM show almost perfect results (column d,e). As we analyzed, we think the main reason is that NMT's attention machanism and NTM's memory component help.
Then we wonder if the supporting facts input is not perfect, for example if the search step can only mark some facts as supporting facts with high probability. Therefore in this experiment, we input the complete passage including non-supporting facts, and use markers to annotate the begin and end of the supporting facts in the passage. Including non-supporting facts in input brings noise and it requires the model to memorize relevant information and ignore non-relevant ones. From the results in group (iii) we see that NMT and NTM both perform very good with just a little expected drop from using supporting facts only. NTM is shown to be better in that NTM's explicit memory component and content addressing mechanism could directly target the relevant information stored.

Although NMT and NTM showed good capability of solving the generation step in experiments above. supporting facts still need to be identified before the models could be applied. As we analyzed above, MemNN-S is good at searching supporting facts. Thus, we use MemNN-S to generate facts for those models and then apply NMT based on its fact-searching results.
We see this combination (column h) improves the average performance over baseline (column a). It proves that this novel architectural combination fusees the advantage of each, i.e. the memory component and attention mechanism.

One major advantage of applying DL models to solve QA is that they can be learned/tuned end-to-end. Therefore the combination of two models through separate training may be less advantageous than single model that has both key elements. As we analyzed, NMT's architecture is essentially an attention mechanism over bidirectional RNN which has some memory functionality. Thus we wonder how NMT single model would perform compared to architectural combination shown above. In this group of experiments, we use NMT model for end-to-end QA without the medium step of finding supporting facts to compare with previous architectural combination. That being said, supporting facts are not marked up in both training and testing which reduced a lot information that model can use to learn. So it requires that attention on supporting facts need to be learned as an implicit byproduct by the NMT attention mechanism. We run NMT on both small and large training sets. The experiment results (column i,j) show that it got 72\% without any tuning or specialization on QA problem. This result on small training set is within a reasonable gap to previous architectural combination, considering the model does not use supporting facts at all. Furthermore, when training data is sufficient, the accuracy is even comparable to the state-of-the-art accuracy (92\%) from \cite{endtoend} where specialized features are added and tuned specifically for this data. In sum, we think the NMT has potential to address certain QA problem, as it has both memory supported in RNN and the attention over the memory.

\section{Conclusions}
We studied several state-of-the-art DL models proposed for different learning tasks on solving QA problem. Experiment results suggest that a good agent need to remember and forget some facts when necessary and external memory may be a choice. We are also convinced that to generate answer, appropriate attention mechanism can do well. Therefore, we believe a DL model combining memory and attention mechanism great potential on handling QA problem.

%
\bibliographystyle{abbrv}
\bibliography{reference}  

\end{document}